\documentclass[sigconf]{acmart}

\AtBeginDocument{%
  \providecommand\BibTeX{{%
    \normalfont B\kern-0.5em{\scshape i\kern-0.25em b}\kern-0.8em\TeX}}}

\usepackage[misc]{ifsym}
\usepackage{bbm}
\usepackage{pifont}
\usepackage{bm}
\usepackage{multirow}
\usepackage{enumitem}
\usepackage{hyperref}
\hypersetup{
    colorlinks=true,
    linkcolor=blue,
    filecolor=magenta,      
    urlcolor=blue
    }
\usepackage{caption}
\usepackage{subcaption}
\usepackage[linesnumbered,ruled,vlined]{algorithm2e}

\newenvironment{myquote}[1]%
  {\list{}{\leftmargin=#1\rightmargin=#1}\item[]}%
  {\endlist}

{\noindent\hangindent=6pt\hangafter=1}

\newcommand{\divea}{\textsf{D\lowercase{iv}EA }}
\newtheorem{definition}{Definition}

\usepackage{tikz}
\usepackage{xcolor}

\SetCommentSty{mycommfont}

\copyrightyear{2022} 
\acmYear{2022} 
\setcopyright{acmlicensed}\acmConference[CIKM '22]{Proceedings of the 31st ACM International Conference on Information and Knowledge Management}{October 17--21, 2022}{Atlanta, GA, USA}
\acmBooktitle{Proceedings of the 31st ACM International Conference on Information and Knowledge Management (CIKM '22), October 17--21, 2022, Atlanta, GA, USA}
\acmPrice{15.00}
\acmDOI{10.1145/3511808.3557352}
\acmISBN{978-1-4503-9236-5/22/10}

\begin{document}

\title[High-quality Task Division for Large-scale Entity Alignment]{High-quality Task Division for Large-scale Entity Alignment}

% \author{Bing Liu$^\text{\Letter}$, Wen Hua, Guido Zuccon}
% % \authornote{Both authors contributed equally to this research.}
% \email{{bing.liu,w.hua,g.zuccon}@uq.edu.au}
% % \email{bing.liu@uq.edu.au}
% % \orcid{1234-5678-9012}
% % \author{Wen Hua}
% % \email{w.hua@uq.edu.au}
% \affiliation{%
%   \institution{The University of Queensland}
%   \streetaddress{St. Lucia}
%   \city{Brisbane}
%   \state{QLD}
%   \country{Australia}
%   \postcode{4067}
% }

% \author{Genghong Zhao$^1$, Xia Zhang$^2$}
% \email{{zhaogenghong,zhangx}@neusoft.com}
% \affiliation{%
%  \institution{$^1$Neusoft Corporation, $^2$Neusoft Research of Intelligent Healthcare Technology, Co. Ltd.}
%  \streetaddress{}
%  \city{Shenyang}
%  \state{Liaoning}
%  \country{China}
%  \postcode{}
% }

\author{Bing Liu}
% \authornote{Corresponding author.}
\email{bing.liu@uq.edu.au}
\affiliation{%
  \institution{The University of Queensland}
  \streetaddress{}
  \city{}
  \state{}
  \country{}
  \postcode{}
}
\author{Wen Hua}
\email{w.hua@uq.edu.au}
\affiliation{%
  \institution{The University of Queensland}
  \streetaddress{}
  \city{}
  \state{}
  \country{}
  \postcode{}
}
\author{Guido Zuccon}
\email{g.zuccon@uq.edu.au}
\affiliation{%
  \institution{The University of Queensland}
  \streetaddress{}
  \city{}
  \state{}
  \country{}
  \postcode{}
}
\author{Genghong Zhao}
\email{zhaogenghong@neusoft.com}
\affiliation{%
 \institution{Neusoft Research of Intelligent Healthcare Technology, Co. Ltd.}
 \streetaddress{}
 \city{}
 \state{}
 \country{}
 \postcode{}
}
\author{Xia Zhang}
\email{zhangx@neusoft.com}
\affiliation{%
  \institution{Neusoft Corporation}
  \streetaddress{}
  \city{}
  \state{}
  \country{}
  \postcode{}
}

% \def \authors{Bing Liu, Wen Hua, Guido Zuccon, Genghong Zhao, Xia Zhang}


\begin{abstract}
Entity Alignment (EA) aims to match equivalent entities that refer to the same real-world objects and is a key step for Knowledge Graph (KG) fusion. Most neural EA models cannot be applied to large-scale real-life KGs due to their excessive consumption of GPU memory and time.
One promising solution is to divide a large EA task into several subtasks such that each subtask only needs to match two small subgraphs of the original KGs. However, it is challenging to divide the EA task without losing effectiveness. Existing methods display low coverage of potential mappings, insufficient evidence in context graphs, and largely differing subtask sizes.

In this work, we design the \divea framework for large-scale EA with high-quality task division. 
To include in the EA subtasks a high proportion of the potential mappings originally present in the large EA task, we devise a counterpart discovery method that exploits the locality principle of the EA task and the power of trained EA models. Unique to our counterpart discovery method is the explicit modelling of the chance of a potential mapping.
We also introduce an evidence passing mechanism to quantify the informativeness of context entities and find the most informative context graphs with flexible control of the subtask size.
Extensive experiments show that \divea achieves higher EA  performance than alternative state-of-the-art solutions.
\end{abstract}





\maketitle

\section{Introduction}

% background of KG and EA
A Knowledge Graph (KG) stores entities and their relationships in a graph form. KGs have been widely used to model knowledge within many downstream applications in recommender systems~\cite{DBLP:journals/corr/abs-2003-00911}, natural language processing~\cite{DBLP:journals/csur/SmirnovaC19,DBLP:journals/bib/FeiRZJL21}, question answering~\cite{DBLP:conf/ijcai/LanHJ0ZW21}, among others~\cite{DBLP:journals/tnn/JiPCMY22}.
A single KG is usually far from being complete\footnote{Though a few large-scale KGs exist , e.g., Wikidata~\cite{DBLP:journals/cacm/VrandecicK14}, Yago~\cite{DBLP:conf/www/SuchanekKW07}, DBPedia~\cite{DBLP:conf/semweb/AuerBKLCI07}.}: the use of a largely incomplete KG might fail to support the downstream application that relies on it.
Fusing different existing KGs is a promising way of deriving a richer, more comprehensive one that ultimately can better support the downstream tasks that employ the KG. Entity Alignment (EA) aims to identify entities that refer to the same real-world objects in different KGs: For each entity in the \textit{source} KG, EA aims to identify its \textit{counterpart} from the \textit{target} KG. Entity Alignment is a key step for KG fusion.

State-of-the-art EA models typically use Graph Convolutional Networks (GCNs) to encode KGs entities  into embeddings, and then match entities in the vector space~\cite{DBLP:conf/ijcai/WuLF0Y019,DBLP:conf/cikm/MaoWXWL20,DBLP:conf/emnlp/LiuCPLC20,DBLP:conf/aaai/00020WD21}.
In the training stage, some \textit{seed mappings} (i.e. pre-aligned equivalent entity pairs) are used to learn the embeddings of unmatched entities; these embeddings are then used to predict unknown mappings.
The pre-aligned entities, also called \textit{anchor} entities, provide \textit{evidence} for recognizing unknown mappings.
Though existing neural EA models have achieved promising performance, two key problems arise when these models are used with large-scale KGs, i.e. KGs with millions or billions of entities. Problem 1: GPU memory -- storing all entity embeddings requires more memory than what is available on the computer infrastructure undertaking the task, easily causing GPUs (which often have a more limited memory than CPU systems) to raise Out-of-Memory exceptions and errors. Problem 2: efficiency --  the time required for performing the EA task on large KGs is infeasible, despite running these on powerful GPU infrastructure.

% Motivation of task division
One promising solution to these problems is dividing a large EA task into several small and self-contained subtasks~\cite{DBLP:journals/pvldb/GeLCZG21,zeng2022entity}. Each subtask only consists of two small subgraphs produced from the original KGs so that it fits a limited amount of memory. Existing neural EA models can be applied to each subtask to detect unknown mappings according to the included seed mappings, and the mappings found in all subtasks jointly form the final alignment results. Besides solving the memory issue, such a solution also naturally provides the opportunity to further speed up the EA task through parallel processing. Nevertheless, task division inevitably degrades the overall alignment effectiveness, and notable challenges arise to mitigate this side effect.

%How to guarantee the recall of entity alignment?
\textbf{Challenge 1: Subtasks with high coverage of potential mappings.} The subtasks obtained by dividing a large EA task are processed independently from each other (allowing extreme parallelisation).
It is then crucial that equivalent entities are allocated to the same subtask -- that is, the task division needs to guarantee a high coverage of the potential mappings in the subtasks.
The CPS algorithm attempts to achieve this by partitioning the source and target KGs collaboratively~\cite{DBLP:journals/pvldb/GeLCZG21}. It does so by first dividing the source KG using a widely-used graph partition method (METIS)~\cite{karypis1997metis}.
For each source subgraph, it then collects all the anchor entities and adjusts the edge weights of the target KG based on the corresponding anchors (increasing the weights of edges connected to the anchors and decreasing the others).
It then partitions the weight-adjusted target KG using the same METIS algorithm. By doing so, most of the seed mappings are expected to be assigned to the same subtask since METIS aims to minimise the sum of the weights of the cut edges.
CPS has been extended by further enhancing the coverage of seed mappings in each subtask using a bidirectional graph partition strategy (SBP) and iterative re-partition with augmented seed mappings generated by the trained EA model (I-SBP)~\cite{zeng2022entity}.
In these methods, for the unmatched entities in each source subgraph, the likelihood that entities in the chosen target subgraph are the ground-truth counterparts is never modelled directly. As a result, the coverage of potential mappings remains to improve.

%High precision in entity alignment
%How to guarantee the accuracy of entity alignment?
\textbf{Challenge 2: High effectiveness in Entity Alignment.} A subtask should contain sufficient \textit{evidence}, such as the KG structure and the supervision signals from seed mappings, in order to achieve high alignment effectiveness. This motivates the existing methods  CPS~\cite{DBLP:journals/pvldb/GeLCZG21}, SBP and I-SBP~\cite{zeng2022entity} to adopt METIS for graph partition, potentially minimising edge-cuts (or structural loss). For a set of unmatched entities from one KG, we call the entities that provide alignment evidence as \textit{context entities}, and the subgraph containing unmatched entities and context entities is called \textit{context graph}. However, KG structure is not the only evidence that can guide entity alignment. Without a clear understanding of the underlying evidence passing mechanism in existing neural EA models, METIS only uses an indirect optimisation of the alignment effectiveness and hence the derived context graph is not sufficiently informative.

%How to balance the subtask size?
\textbf{Challenge 3: Control subtask size.} Controlling and often balancing the subtask sizes is important for guaranteeing the overall performance of the EA task, especially when the subtasks are conducted independently and in parallel. However, the existing algorithms produce subtasks of largely varied sizes.
For example, in our preliminary study, we divide a large-scale EA task (each KG has over 2 million entities) into 200 subtasks with CPS and I-SBP. CPS produces subtasks with sizes varying from 22,032 to 24,075 entities, while I-SBP displays a larger size variance, with sizes ranging from 28,442 to 46,570 entities. This calls for a better solution that can flexibly control the subtask size.

To address these challenges, we propose \divea to align entities between large-scale KGs with a better task division. Our framework consists of three main modules:
(1) Unmatched source entity division, which aims to divide the unmatched source entities into multiple coherent groups;
(2) Counterpart discovery, which selects counterpart candidates for each group of unmatched source entities so that the coverage of potential mappings can be improved;
(3) Context building, which constructs informative context graphs for a group of unmatched source entities and their counterpart candidates. The two context graphs in each subtask are then fed into existing neural EA models for entity alignment.
\divea runs in an iterative manner where the new mappings generated by the trained EA model are reused to enhance the previous modules.

Overall, our main contributions are:
% \vspace{-10pt}
\begin{itemize}[leftmargin=*]
    \item We propose the \divea framework to support entity alignment in large-scale knowledge graphs with high-quality task division.
    \item We propose a progressive counterpart discovery method, which combines the locality principle of the EA task and the power of trained EA models, to guarantee high coverage of potential mappings within subtasks.
    \item We design an evidence passing mechanism to quantify the informativeness of context graphs, and build informative context graphs given a certain size budget. This naturally enables flexible control of the subtask size.
    \item We provide extensive experiments demonstrating the significant improvements of our \divea framework compared to state-of-the-art baselines for building subtasks for neural EA models.
\end{itemize}

\section{Related Work}

\subsection*{Neural Entity Alignment}

Many existing works have studied Entity Alignment. The traditional works~\cite{DBLP:conf/semweb/FariaPSCC14,DBLP:journals/pvldb/SuchanekAS11,DBLP:conf/semweb/Jimenez-RuizG11} compare the attribute values of entities with heuristic methods and apply reasoning rules to infer potential mappings. One drawback of these methods is that they depend on domain-specific experts' efforts.
The success of deep learning~\cite{DBLP:journals/nature/LeCunBH15} in  Computer vision and Natural Language Processing inspired the research of neural EA methods. EA methods based on different neural networks have become the mainstream and achieved significant progress~\cite{DBLP:conf/emnlp/WangLLZ18,DBLP:conf/ijcai/WuLF0Y019,DBLP:conf/cikm/MaoWXWL20,DBLP:conf/wsdm/MaoWXLW20,DBLP:journals/pvldb/SunZHWCAL20,9174835}. In addition to various models, different learning paradigms~\cite{DBLP:conf/ijcai/SunHZQ18,DBLP:conf/emnlp/LiuSZHZ21,DBLP:conf/ijcai/ChenTCSZ18,DBLP:conf/emnlp/ZhangZ0WWHDC021,DBLP:conf/aaai/XinSHLHQ022} are also explored to train neural EA models more effectively. Unlike reasoning-based methods, neural EA models can automatically learn how to match entities from seed mappings. Also, they have the potential to deal with sparse KGs without much attribute information.
A general idea of neural EA models is: encoding the entities into vector embeddings with KG encoder, and then match entities based on their distance in the vector space. Typically, the neural EA models are trained on GPU devices, which usually have limited GPU memory.
Though state-of-the-art EA models have achieved promising effectiveness, most neural models can only run on small datasets, and cannot be used to match large-scale KGs~\cite{DBLP:journals/pvldb/GeLCZG21,zeng2022entity}. This issue is critical for the application of neural EA models in a real-world scenario. Recently, the research community has begun exploring dividing a big EA task to small subtasks to solve this problem. Two previous works propose to partition large-scale KGs into small subgraphs and then pair the source and target subgraphs to form many small subtasks~\cite{DBLP:journals/pvldb/GeLCZG21,zeng2022entity}.
These frameworks have obvious limitations in EA effectiveness and controlling subtask size. We propose a new framework for EA division to overcome these limitations.

\subsection*{Graph Convolutional Networks}

Graph Convolutional Networks (GCNs)~\cite{DBLP:conf/iclr/KipfW17,DBLP:conf/iclr/VelickovicCCRLB18,DBLP:journals/aiopen/ZhouCHZYLWLS20,DBLP:journals/tnn/WuPCLZY21} are now very popular in extracting the features of graphs. In EA areas, they have been the mainstream method for encoding KGs. Most SOTA EA models~\cite{DBLP:conf/ijcai/WuLF0Y019,DBLP:conf/cikm/MaoWXWL20,DBLP:conf/emnlp/LiuCPLC20,DBLP:conf/aaai/00020WD21} use GCNs as KG encoders.
We analyse the effect of dropped entities from KGs on the training and inference processes of GCN encoder. It inspires us how to quantify the informativeness of context graphs.
Message Passing Mechanism is widely used to formulate different GCN variants~\cite{DBLP:conf/icml/GilmerSRVD17}. We simulate the message passing mechanism to formulate our evidence passing mechanism, which is specific to the EA division.

\section{Problem Formulation}
A Knowledge Graph $\mathcal{G}=(E,R,T)$ typically consists of a set of entities $E$, a set of relations $R$, and a set of triples $T \in E \times R \times E$.
Given one source KG $\mathcal{G}^s$, one target KG $\mathcal{G}^t$, and some pre-aligned seed mappings $M^l=\{ (e^s \in E^s, e^t \in E^t) \}$, Entity Alignment aims to identify potential mappings $M^p = \{ (e^s \in E^s, e^t \in E^t) \}$. The source entities and target entities of seed mappings $M^l$ are also called anchor entities and denoted as $E^{s,a}$ and $E^{t,a}$, while the other unmatched entities are denoted as $E^{s,u}$ and $E^{t,u}$.
The goal of \textit{dividing the EA task} $\langle \mathcal{G}^s, \mathcal{G}^t, {M}^l \rangle$ is to formulate $N$ subtasks $\langle \widetilde{\mathcal{G}}^s_i \subset \mathcal{G}^s, \widetilde{\mathcal{G}}^t_i \subset \mathcal{G}^t, \widetilde{M}^l_i \subset {M}^l \rangle, i=1...N$, under the size limitation constraint $|\widetilde{E}^s_i|+|\widetilde{E}^t_i| \leq S$.
The outputted mappings $\widetilde{M}^p_i$ from each subtask jointly form the final predicted mappings, i.e. $M^p = \bigcup_{i=1...N} \widetilde{M}^p_i$.

The overall objective of dividing the EA task is expressed by Eq.~\ref{eq:overall_objective}, where $\mathsf{EA}(\cdot)$ is the employed neural EA model, and $\mathsf{Eval}(\cdot)$  evaluates the effectiveness of EA:
\begin{equation}
    \arg\max_{\{(\widetilde{\mathcal{G}}^s_i, \widetilde{\mathcal{G}}^t_i)\}, i=1...N} \mathsf{Eval}\left( \bigcup_{i=1...N} \mathsf{EA}(\widetilde{\mathcal{G}}^s_i, \widetilde{\mathcal{G}}^t_i, \widetilde{M}^l_i) \right)
    \label{eq:overall_objective}
\end{equation}

\noindent
To solve the overall objective, we assume the unmatched source entities are divided using an existing graph partition method, and we focus on solving the following two sub-objectives:

\noindent
\textit{Sub-objective 1}: Given a set of unmatched source entities $\widetilde{E}^{s,u}$, select counterpart candidates $\widetilde{E}^{t,u,*} = \mathsf{selectCandidates}(\widetilde{E}^{s,u})$ so that the recall of ground-truth counterparts is maximised:
\begin{equation}
    \widetilde{E}^{t,u,*} = \arg\max_{\widetilde{E}^{t,u}} \mathsf{Recall}(\widetilde{E}^{s,u}, \widetilde{E}^{t,u})
    \label{eq:objective1}
\end{equation}

\noindent
\textit{Sub-objective 2}: Given the unmatched source entities $\widetilde{E}^{s,u}$ and their counterpart candidates $\widetilde{E}^{t,u}$, build their context graphs  $\widetilde{\mathcal{G}}^{s,*}, \widetilde{\mathcal{G}}^{t,*} = \mathsf{buildContext}(\widetilde{E}^{s,u}, \widetilde{E}^{t,u})$ so that they can receive the most evidence for matching.
Formally, this is formulated as in Eq.~\ref{eq:objective2}, where $\widetilde{E}^{s}$ and $\widetilde{E}^{t}$ are the entity set of subgraphs $\widetilde{\mathcal{G}}^{s}$ and $\widetilde{\mathcal{G}}^{t}$ respectively:
\begin{equation}
    \widetilde{\mathcal{G}}^{s,*}, \widetilde{\mathcal{G}}^{t,*} = \arg\max_{\widetilde{\mathcal{G}}^s, \widetilde{\mathcal{G}}^t} \mathsf{Eval}\left(\mathsf{EA}(\widetilde{\mathcal{G}}^s, \widetilde{\mathcal{G}}^t, \widetilde{M}^l )\right)
    \label{eq:objective2}
\end{equation}
$$s.t. \text{ } \widetilde{E}^{s,u} \subset \widetilde{E}^s, \widetilde{E}^{t,u} \subset \widetilde{E}^t$$

\section{The \divea Framework}

\subsection{Overview}

\begin{figure}
    \centering
    \includegraphics[width=8cm]{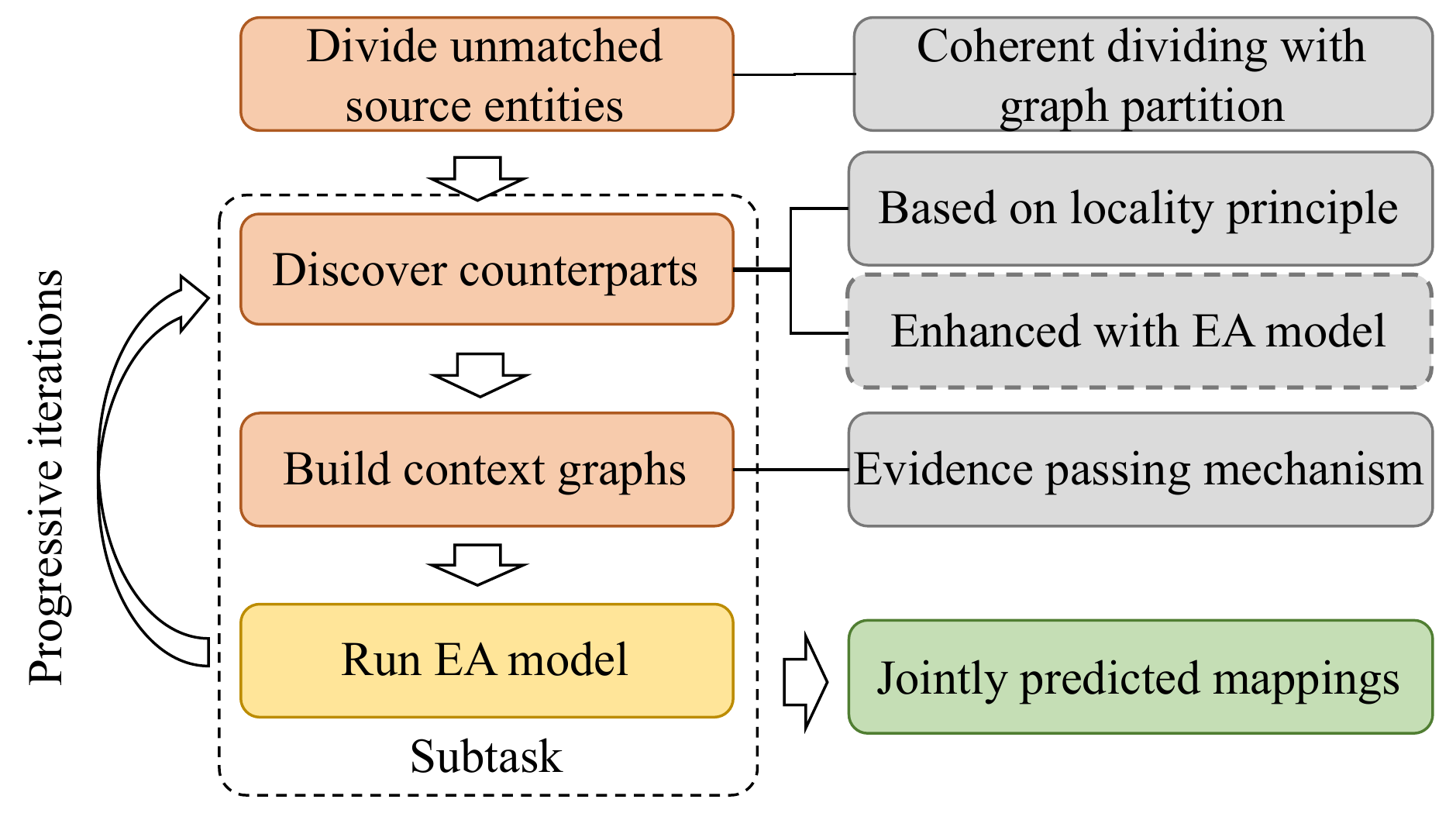}
    \vspace{-8pt}
    \caption{Overview of the proposed \divea framework.}
    \label{fig:overview_prog}
    \vspace{-12pt}
\end{figure}

An overview of our \divea framework is given in Fig.~\ref{fig:overview_prog}. \divea consists of three modules -- (1) division of unmatched source entities, (2) counterpart discovery, and (3) context builder.
The whole process starts with dividing unmatched source entities.
We divide the original source KG $\mathcal{G}^s$ into $N$ coherent partitions $\{\mathcal{P}_i, i=1...N\}$ with \textit{METIS}~\cite{karypis1997metis}, which is a mini-cut based graph partition tool and can derive subgraphs with relatively balanced (i.e. similar) sizes.
We only collect the unmatched entities in each partition $\mathcal{P}_i$ to form $\widetilde{E}^{s,u}_i$.
Then, a subtask is run in an iterative way for each group of unmatched source entities $\widetilde{E}^{s,u}_i$.
Within each iteration, we firstly look for the counterparts $\widetilde{E}^{t,u}_i$ of $\widetilde{E}^{s,u}_i$. Our counterpart discovery is based on the locality principle of EA. Besides, it is enhanced with entity similarities provided by the EA model when available.
Then, we build the context graphs $\widetilde{\mathcal{G}}^s_i, \widetilde{\mathcal{G}}^t_i$ for $\widetilde{E}^{s,u}_i$ and $\widetilde{E}^{t,u}_i$.
We propose one evidence passing mechanism to quantify the informativeness of a certain subgraph w.r.t. a set of unmatched entities, either $\widetilde{E}^{s,u}_i$ or $\widetilde{E}^{t,u}_i$. With such a mechanism, we search for the most informative subgraph as the context graph.
At the end of each iteration, we train the EA model on the context graphs.
New (i.e. pseudo) mappings are then inferred and added to the set of seed mappings. Also, the EA model will assist counterpart discovery in the next iteration.
Eventually, the whole process ends, outputting the jointly predicted mappings produced by the neural EA models.

\vspace{-2pt}
\subsection{Counterpart Discovery}\label{sec:select_candidate}

In this section, we examine locality-based counterpart discovery and then explain how this is improved with an EA model.

\subsubsection{Locality-based Counterpart Discovery}

\begin{myquote}{8pt}
    %\begin{definition}
    (Principle of locality). If entity $e^s$ is mapped to $e^t$, the entities semantically related to $e^s$ are likely to be mapped to those semantically related to $e^t$.
%    \end{definition}
\end{myquote}

\noindent
The principle of locality has been used with success in previous work to discover potential mappings according to existing mappings~\cite{DBLP:conf/semweb/Jimenez-RuizG11}.
Our scenario is however different. Given unmatched source entities $\widetilde{E}^{s,u}_i$ from partition $\mathcal{P}_i$, we first collect the seed mappings with source anchors contained in $\mathcal{P}_i$. Then, we assign a weight to the neighbouring entities $e^t$ of the target anchors $\widetilde{E}^{t,a}_i$ as in Eq.~\ref{eq:w_dist}, where $\mathrm{dist}(\cdot)$ means the length of the shortest path from $e^t$ to any entity in $\widetilde{E}^{t,a}_i$.
\begin{equation}
    W^{loc}(e^t) = - \mathrm{dist}(e^t, \widetilde{E}^{t,a}_i)
    \label{eq:w_dist}
\end{equation}

\noindent
The intuition is that entities closer (i.e. with stronger relatedness) to $\widetilde{E}^{t,a}_i$ have higher chances to be the counterparts of $\widetilde{E}^{s,a}_i$.

The coherence of each partition $\mathcal{P}_i$ is critical to performing locality-based counterpart discovery.
Higher coherence between the source anchors and unmatched source entities can increase the chance of finding ground-truth counterparts.
Besides, the coherence between unmatched source entities themselves allows us to take the strategy of progressive counterpart discovery: finding partial mappings firstly and then using them to assist seeking counterparts of the remaining unmatched entities, as described next.

\subsubsection{Enhancing Counterpart Discovery with an EA Model.}
Our framework relies on the use of an EA model within the counterpart discovery process for two reasons.

The first reason is that the use of the EA model can help overcome the sparsity issue associated with seed mappings: having denser mappings is key for locality-based counterpart discovery.
We follow the method used in previous EA methods to generate the pseudo mappings~\cite{DBLP:conf/wsdm/MaoWXLW20,DBLP:conf/cikm/MaoWXWL20}: the pair $(e^s,e^t)$ will be added to the seed mappings if and only if two entities $e^s$ and $e^t$ are mutually the closest counterpart candidate of each other.
With the enriched seed mappings, the remaining unmatched entities can receive more signals through the locality principle.

The second reason is to introduce additional signals, other than locality, for identifying counterparts. Ideally, the candidates with a high likelihood of being ground-truth counterparts would be kept among the candidates in the subsequent iterations. However, the use locality alone cannot guarantee this.
Therefore, we develop a weight feature to indicate the likelihood that each entity $e^t$ is the true counterpart of a certain source entity $e^s \in \widetilde{E}^{s,u}$.
In particular, we accumulate the top $K$ similarity scores for $e^t$ (Eq.~\ref{eq:f_sim})
, and then use a min-max scaler to normalise the features (Eq.~\ref{eq:f_sim_norm}), with $\alpha$ being a threshold parameter.

\vspace{-6pt}
\begin{equation}
    W'(e^t) = \mathrm{sum} \left(\mathrm{top}(K, \mathrm{sim}(\widetilde{E}^{s,u}, e^t)) \right)
    \label{eq:f_sim}
\end{equation}

\begin{equation}
    W^{sim}(e^t) = \frac{W'(e^t)- \min_{e\in \widetilde{E}^{t,u}} W'(e)}{\max_{e\in \widetilde{E}^{t,u}} W'(e) - \min_{e\in \widetilde{E}^{t,u}} W'(e) } - \alpha
    \label{eq:f_sim_norm}
\end{equation}

\noindent
Specially, a weight of 0 is assigned to the entities that have never been selected as candidates, while the entities selected multiple times are given their latest weights.
Finally, we linearly combine the locality-based and similarity-based weights (Eq.~\ref{eq:weight_counterpart}). Note that in the first iteration, only $W^{loc}(e^t)$ is used as weight since the EA model is not trained yet, and as a result, $W^{sim}(e^t)$ cannot be computed.
\begin{equation}
    W(e^t) = W^{loc}(e^t) + \beta \cdot W^{sim}(e^t)
    \label{eq:weight_counterpart}
\end{equation}

\subsubsection{Incorporating Attributes}

Ge et al.~\cite{DBLP:journals/pvldb/GeLCZG21} generate pseudo mappings based on the attribute information, i.e. entity name, to augment the seed mappings before performing EA division. This strategy can naturally be integrated into our method.

\subsection{Building Context Graphs}\label{sec:build_ctx}

Given $\widetilde{E}^{s,u}_i$ and $\widetilde{E}^{t,u}_i$, our goal is to build the corresponding source context graph and target context graph. Once these are obtained, by executing the neural EA model on the context graphs, the potential mappings contained in $\widetilde{E}^{s,u}_i \times \widetilde{E}^{t,u}_i$ can be found.
To this end, we first try to understand the effect of dropped entities on the GCN-based KG encoder in both training and inference procedures. Then, we design an evidence passing mechanism to quantify such effect (i.e. loss of evidence) on a set of unmatched entities, either $\widetilde{E}^{s,u}_i$ or $\widetilde{E}^{t,u}_i$.
Eventually, we drop the entities from the original KG that have the least effect when deriving the context graphs.

\subsubsection{Understanding the Effect of Dropped Entities}

\begin{figure}
    \centering
    \includegraphics[width=8.5cm]{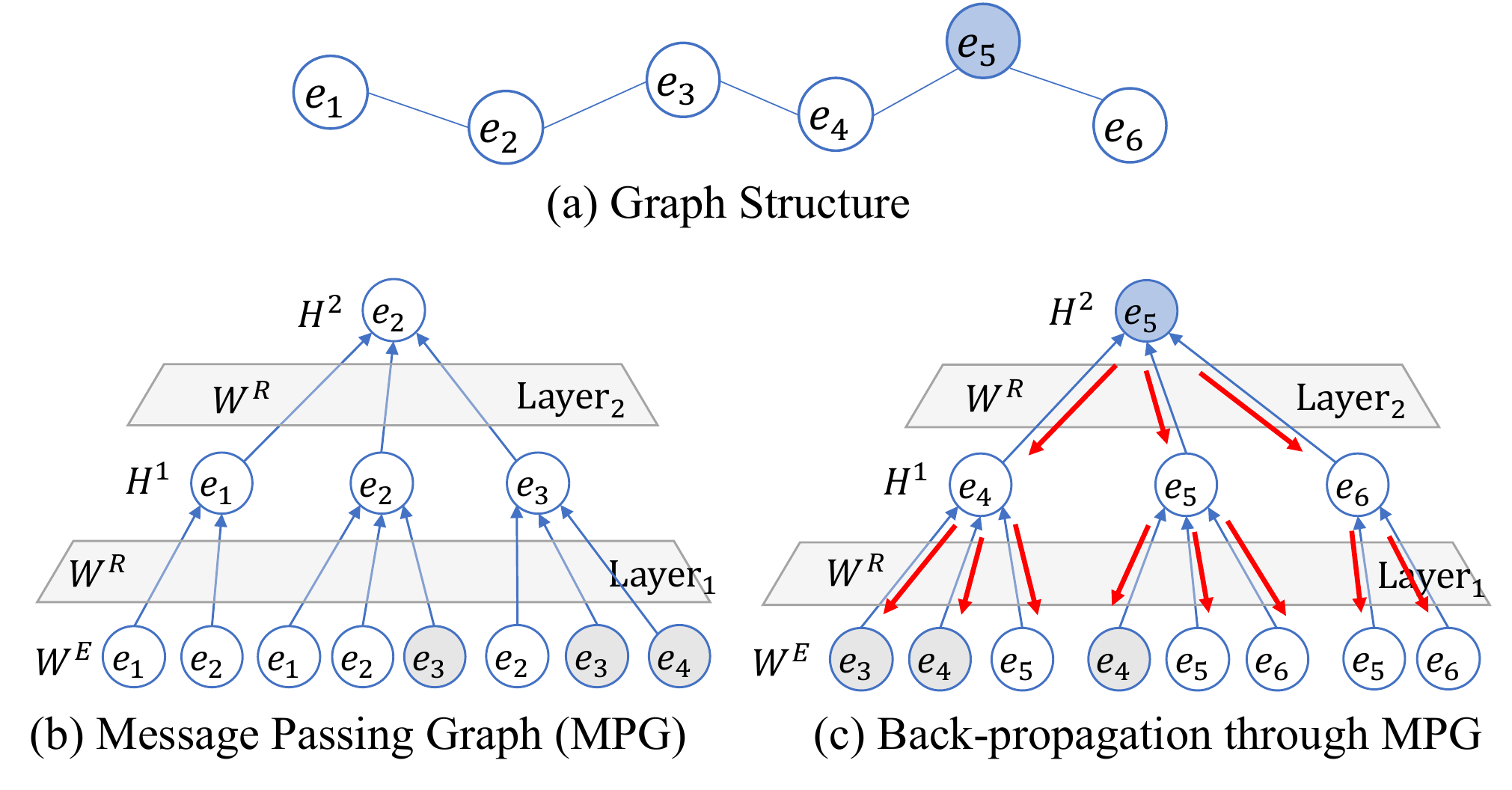}
    % \vspace{-20pt}
    \caption{Message passing graph.}
    \label{fig:gcn_mpg}
    % \vspace{-10pt}
\end{figure}

\begin{definition}
    (Message Passing Graph). The Message Passing Graph (MPG) of an entity $e$ is a hierarchical graph of depth $L$ describing how neural messages are sent to it from its $L$-hop neighbours in an $L$-layer GCN model. Each internal node has all neighbouring entities as well as itself as its children pointing to it. The paths connecting the leaves to the root are called \textit{Message Paths}.
\end{definition}

\noindent
Some simple examples of MPGs are shown in Fig.~\ref{fig:gcn_mpg}.
%In Fig.~\ref{fig:gcn_mpg} we show toy examples of MPG.
Suppose one GCN model is run on the graph with anchor $e_5$ in Fig.~\ref{fig:gcn_mpg} (a).  The hierarchical graphs in Fig.~\ref{fig:gcn_mpg} (b) and (c) are the MPGs of $e_2$ and $e_5$.
In Fig.~\ref{fig:gcn_mpg} (b), we show that entity $e_2$ gets its representation by flowing the neural messages from the leaf nodes to the root node in its MPG
in a forward-propagation process. The representation is only determined by the message paths and related parameters in the MPG.
If one entity is dropped, the message paths containing it will be cut. For example, in Fig.~\ref{fig:gcn_mpg} (b), dropping $e_1$ will cause three message paths involving it to be broken up. As a result, the representation of $e_2$ is affected.

\noindent
In Fig.~\ref{fig:gcn_mpg} (c), we show one anchor entity $e_5$ can back-propagate gradient messages to the leaf nodes in its MGP in the inverse direction of each message path. The parameters related to each path will be updated.
If one entity occurring in the MPG of one anchor is dropped, the message paths containing it will be interrupted and the parameters related to these message paths will be affected during training. This will eventually affect some entities' representations in the inference stage.
For example, in Fig~\ref{fig:gcn_mpg} (c), dropping $e_4$ will affect the parameters for $e_3$ and $e_4$ and eventually indirectly affect the representation of $e_2$.

In general, one entity may occur in multiple MPGs of different entities. Dropped entities affect both forward-propagation and back-propagation processes. Eventually, the representations of unmatched entities worsen and the EA performance decreases.

\subsubsection{Evidence Passing Mechanism for Quantifying Informativeness of Context Graph.}
%\vspace{1pt}

Given a whole KG $\mathcal{G}$ and a set of unmatched entities $\widetilde{E}^{u}_i$, our goal is to quantify the informativeness of a certain context graph $\mathcal{G}' \subset \mathcal{G}$ w.r.t. $\widetilde{E}^{u}_i$. The entity set and anchor set of $\mathcal{G}'$ are denoted as $E'$ and $E'^a$.

In the training stage, the evidence flows from the anchors to the parameters. We initialise $h^0_e=\mathbbm{1}_{e \in E'^a}$ for $e \in E'$.
The evidence model has the same number of layers $L$ as the GCN encoder for EA.
Each layer is defined as in Eq.~\ref{eq:msg_passing}, where $h^l_e$ is the node representation of $e$ at layer $l$, $m^l_{e' \rightarrow e}$ is the message sent from $e'$ to $e$, and $\mathcal{N}_e$ denotes the neighbours of $e$. The equation represents this process: each entity sends a message to its neighbours (first item), and then each entity aggregates the received messages (second item).

\begin{equation}
    m^{l+1}_{e' \rightarrow e} = h^{l}_{e'}, h^{l+1}_e = \sum_{e' \in \{e\} \cup \mathcal{N}_e} m^{l+1}_{e' \rightarrow e}
    \label{eq:msg_passing}
\end{equation}

\noindent
Through the $L$ stacked layers of the evidence model, we obtain the amount of evidence $h^L_e$ each entity $e$ can receive.
Then, we normalise it with Eq.~\ref{eq:evi_update}. Here,
$(h^{l+1}_e)_{\mathcal{G}}$ denotes the total messages entity $e$ can receive from the whole KG $\mathcal{G}$, which is a constant and will be further explained later. The function $2 \cdot \mathrm{sigmoid}(x)-1$ transforms $x \geq 0$ into $[0,1)$ and is more sensitive to smaller values.

\begin{equation}
    h^{o, L}_e = 2 \cdot \mathrm{sigmoid} \left(\lambda \cdot \frac{ h^{L}_e } { (h^{L}_e)_{\mathcal{G}} } \right) - 1
    \label{eq:evi_update}
\end{equation}

In the inference stage, the evidence continues flowing from parameters to the final entity embeddings.
We apply the message passing mechanism defined in Eq.~\ref{eq:msg_passing} with $h^{o, L}_e$ as the input layer, and normalise it with Eq.~\ref{eq:evi_update2}, which is the same transformation as Eq.~\ref{eq:evi_update} but with $h^{2L}_e$ as input.

\begin{equation}
    h^{o, 2L}_e = 2 \cdot \mathrm{sigmoid} \left(\lambda \cdot \frac{ h^{2L}_e } { (h^{2L}_e)_{\mathcal{G}} } \right) - 1
    \label{eq:evi_update2}
\end{equation}

\noindent
The output $h^{o,2L}_e$ denotes the evidence that entity $e$ eventually receives.
Therefore, the informativeness of a context graph $\widetilde{\mathcal{G}}_i$ w.r.t. a set of unmatched entities $\widetilde{E}^{u}_i$ can be obtained by summing up all the received evidence as in Eq.~\ref{eq:informative}. Here, we use $(h^{o,2L}_e)_{\mathcal{G}'}$ to emphasize $h^{o,2L}_e$ is obtained with ${\mathcal{G}'}$ as the context graph.

\begin{equation}
    I(\mathcal{G}', \widetilde{E}^{u}_i) = \sum_{e \in \widetilde{E}^{u}_i} (h^{o, 2L}_e)_{\mathcal{G}'}
    \label{eq:informative}
\end{equation}

To avoid confusion, we add more considerations on how to compute $(h^{L}_e)_\mathcal{G}$ and $(h^{2L}_e)_\mathcal{G}$, which both are constants and used in Eq.~\ref{eq:evi_update} and Eq.~\ref{eq:evi_update2}.
(1) Since $\mathcal{G} \subset \mathcal{G}$, the process of applying the evidence model on the whole KG $\mathcal{G}$ is exactly the same as on any other context graph $\mathcal{G}' \subset \mathcal{G}$.
(2) In Eq.~\ref{eq:evi_update}, we would have $\frac{(h^{L}_e)_{\mathcal{G}'}}{(h^{L}_e)_{\mathcal{G}'}}=1$. Thus, $h^{o,L}_e$ can be computed as a constant.
(3) Since $\mathcal{G}' \subset \mathcal{G}$, we can obtain the value of $h^L_e$ for any $e \in E'$.

Compared to a general GCN model, the differences with our evidence model can be summarised as: (1) The node representations are scalar instead of vectors; (2) It has no learnable parameters and thus does not need training.
(3) For the reasons above, it can run on very large graphs using common GPUs.

\begin{algorithm}[t!]
    \caption{The \divea Framework}
    \label{alg}
    \SetKwInput{KwInput}{Input}                % Set the Input
    \SetKwInput{KwOutput}{Output}              % set the Output
    \DontPrintSemicolon

    \KwInput{$\mathcal{G}^s$, $\mathcal{G}^t$, $M^{l}$, $N$, $S$}
    \KwOutput{$M^p$, $\{ (\widetilde{\mathcal{G}}^s_i, \widetilde{\mathcal{G}}^t_i),i = 1...N \}$}

    \SetKwFunction{FMain}{Main}
    \SetKwFunction{FSubtask}{Subtask}
    \SetKwFunction{FCandi}{selectCandidates}
    \SetKwFunction{FBuildCtx}{buildContext}
    \SetKwFunction{FMatch}{EA}
    \SetKwFunction{FTrainEA}{TrainEA}
    \SetKwFunction{FInferEA}{InferEA}
    \SetKwFunction{FFilter}{Filter}
    \SetKwFunction{FDivide}{divide}

    \SetKwProg{Fn}{Def}{:}{}
    \Fn{\FMain{}}{
        % Divide unmatched entity set $E^1$ into $\{\widehat{E}^{1}_i, i=1...N\}$ \;
        \tcp{Partition source graph}
        $\{(\mathcal{P}_i,\widetilde{E}^{s,u}_i), i=1...N\}=$ \FDivide{$\mathcal{G}^s$, $N$ } \;
        $M^p = \{\}$ \tcp{Initialize prediction set}
        \tcp{Start the progressive process}
        \For{iterations}{
            \tcp{Run subtasks which can be in parallel}
            $\widetilde{\mathcal{G}}^s_i$, $\widetilde{\mathcal{G}}^t_i$,  $\widetilde{M}^p_i$ = \FSubtask{$\mathcal{P}_i$, $\widetilde{E}^{s,u}_i$, $M^l \cup M^p$, $S$} for $i=1...N$ \;
            $M^p = \bigcup_{i=1...N} \widetilde{M}^p_i$ \tcp{Collect predictions}
        }
        \KwRet{$M^p$, $\{ (\widetilde{\mathcal{G}}^s_i, \widetilde{\mathcal{G}}^t_i), i=1...N \}$}
    }

    \SetKwProg{Fn}{Def}{:}{}
    \Fn{\FSubtask{$\mathcal{P}_i$, $\widetilde{E}^{s,u}_i$, $M$, $S$}}{
        $\widetilde{M}'_i = \{ (e^s, e^t) \in M | e^s \in \mathcal{P}_i \} $ \tcp{Select mappings in $\mathcal{P}_i$}
        $\widetilde{E}^{t,u}_i=$ \FCandi{$\widetilde{E}^{s,u}_i, \widetilde{M}'_i$} \tcp{Use Eq.~\ref{eq:w_dist},\ref{eq:f_sim_norm},\ref{eq:weight_counterpart}}
        \tcp{Update $\widetilde{Sim}_i$ with current EA model when available}
        $\widetilde{Sim}_i = 0$ or updated with \FInferEA($\widetilde{G}^s_{i-1}$, $\widetilde{G}^t_{i-1}$) \;
        \tcp{Build context based on Eq.~\ref{eq:msg_passing},\ref{eq:evi_update},\ref{eq:drop_cost}}
        $\widetilde{\mathcal{G}}^s_i, \widetilde{\mathcal{G}}^t_i, \widetilde{M}^l_i = $ \FBuildCtx{$\widetilde{E}^{s,u}_i$, $\widetilde{E}^{t,u}_i$, $S$, $\widetilde{Sim}_i$} \;
        \tcp{Train EA model and predict new mappings}
        $\widetilde{M}^p_i=$ \FMatch{$\widetilde{\mathcal{G}}^s_i, \widetilde{\mathcal{G}}^t_i, \widetilde{M}^l_i$} \;
        \KwRet $\widetilde{\mathcal{G}}^s_i$, $\widetilde{\mathcal{G}}^t_i$,  $\widetilde{M}^p_i$\;
    }
\end{algorithm}

\subsubsection{Context Entity Selection}

Since the informativeness of context graphs can be quantified, we can formulate the task of building the most informative context graph as a search problem.
However, we choose a simpler alternative way.
%For simplicity, we implement it in a simple way instead.
We estimate the effect (i.e. cost) of dropping an entity $e \in E \setminus \widetilde{E}^{u}_i$ from the whole KG as: % in Eq.~\ref{eq:drop_cost}.

\begin{equation}
    C(e) = I(\mathcal{G}, \widetilde{E}^{u}_i)  - I(\mathcal{G}\setminus \{e\}, \widetilde{E}^{u}_i), e \in E \setminus \widetilde{E}^{u}_i
    \label{eq:drop_cost}
\end{equation}

\noindent
Then, we drop a certain number of entities causing the least cost from $E \setminus \widetilde{E}^{u}_i$. The remaining are context entities $\widetilde{E}^c_i$.
The context graph is then the subgraph that only contains $\widetilde{E}^{u}_i \cup \widetilde{E}^c_i$.

\subsubsection{Overall Process of Building Both Context Graphs}

Both the source and target context graphs are built based on our evidence passing mechanism.
We first build the source context graph $\widetilde{\mathcal{G}}^s_i$ for the unmatched source entities $\widetilde{E}^{s,u}_i$ from the source KG $\mathcal{G}^s$.
Among the context entities, some are source anchors $\widetilde{E}^{s,a}_i$ with corresponding target anchors $\widetilde{E}^{t,a}_i$, while the others connect the anchors with entities in $\widetilde{E}^{s,u}_i$.
Then, when building the target context graph $\widetilde{\mathcal{G}}^t_i$ for $\widetilde{E}^{t,u}_i$, we start by adding the target anchors $\widetilde{E}^{t,a}_i$ as context entities. Then, we treat all other entities $E^t \setminus (\widetilde{E}^{t,a}_i \cup \widetilde{E}^{t,u}_i)$ as connecting entities because the first context graph has determined the valid target anchors $\widetilde{E}^{t,a}_i$ when selecting the source anchors.
Afterwards, we apply the evidence passing mechanism on the target KG $\mathcal{G}^t$, adding some connecting entities into the target context graph $\widetilde{\mathcal{G}}^t_i$.

\subsection{Implementation}

%To make the whole process clear,
We provide the algorithm of \divea in Alg.~\ref{alg}. Also, we provide a few details for implementation:
(1) In line 11 of Alg.~\ref{alg}, only when $\widetilde{E}^{s,u}_i$ or $\widetilde{E}^{t,u}_i$ is changed will we update their respective context graph. We only build the source context graphs once in the first iteration and then cache them.
Conversely, we need to update the target context graphs in each iteration since the counterpart candidates $\widetilde{E}^{t,u}_i$ are updated progressively in our counterpart discovery.
(2) Under the size limitation $|\widetilde{E}^s_i|+|\widetilde{E}^t_i| \leq S$, we can flexibly adjust the sizes of the source and target context graphs, as well as the number of counterpart candidates in the target context graphs to optimise the EA performance.
For example, we can set two factors: the size proportion $\delta_1$ of the source context graph, and the size proportion $\delta_2$ of counterpart candidates in the target context graph.

\section{Evaluation}

We design the experiments from perspectives specific to neural EA division.
(1) We only compare \divea with the existing works focusing on EA division, which is used to serve existing neural EA models but not to replace them. Refer to the early EA division works~\cite{DBLP:journals/pvldb/GeLCZG21,zeng2022entity} if you are interested in the comparison between EA division and neural EA models.
(2) We compare \divea with existing EA division methods under the same subtask number and size limitation. Zeng et al.~\cite{zeng2022entity} compare their method with LargeEA~\cite{DBLP:journals/pvldb/GeLCZG21} only under the same subtask number. Though can achieve better EA effectiveness, their method also builds larger subtasks. It is hard to confirm the source of improvement.

\vspace{-8pt}
\subsection{Comparative Baselines}

We compare \divea with two existing works investigating dividing neural EA tasks. As known, the EA model plays the role of evaluating different division methods. For a fair comparison, we try to train the EA model in a comparable way w.r.t. total training epoch and training mode (i.e. supervised or semi-supervised learning).

\begin{itemize}[leftmargin=*]
    \item CPS variants. Ge et al.~\cite{DBLP:journals/pvldb/GeLCZG21} propose one method called CPS, which partitions the original KGs into two sets of subgraphs, and pair each source subgraph with a target subgraph to form one subtask. In the original work, the EA model is trained in supervised mode. However, our \divea involves semi-supervised training of the EA model. To make a fair comparison, we run the EA model in both supervised and semi-supervised modes and denote two variants as CPS (sup) and CPS (semi) respectively.
    \item SBP variants. Zeng et al.~\cite{zeng2022entity} improve CPS with a bidirectional strategy named SBP. In addition, they develop I-SBP, which trains EA in a semi-supervised manner and enriches the seed mappings with pseudo mappings in each iteration. Like CPS, SBP trains the EA model in a supervised manner. We develop the corresponding semi-supervised version and denote both variants as SBP (sup) and SBP (semi). Note SBP (semi) is different from I-SBP because it does not intervene EA division.
\end{itemize}

\noindent
It is not fair to compare SBP and CPS directly because SBP variants build larger subtasks than CPS. Therefore, \textbf{we compare \divea with CPS and SBP separately}.
In addition, both SBP and CPS build subtasks with varying sizes. We set the subtask size of \divea as their average subtask size when compared with them.

\vspace{-8pt}
\subsection{Metrics}

Following the existing works~\cite{DBLP:journals/pvldb/GeLCZG21,zeng2022entity}, we run two different neural EA models -- RREA~\cite{DBLP:conf/cikm/MaoWXWL20} and GCN-Align~\cite{DBLP:conf/emnlp/WangLLZ18} -- for the built subtasks to measure the performance of EA division methods.
As for the metrics of EA models, we choose Hit@k (k=1,5) and Mean Reciprocal Rank (MRR), which are widely used in the EA community. Hit@k (abbreviated as H@k) indicates the percentage of source entities whose ground-truth counterparts are ranked in the top $k$ positions. MRR assigns each prediction result the score $1/Rank(ground-truth \text{ } counterpart)$ and then averages them.
When reporting the coverage of counterpart discovery, we use the recall of potential mappings, i.e. the proportion covered by the subtasks.
Statistical significance is performed using a paired two-tailed t-test.

\subsection{Datasets and Partition}

We choose five datasets widely used in previous EA research and one dataset specially built for investigating large-scale EA. Each
dataset contains two KGs and a set of pre-aligned entity pairs.
\begin{itemize}[leftmargin=*]
    \item FR-EN, JA-EN, and ZH-EN from DBP15K~\cite{DBLP:conf/emnlp/SunCHWDZ20}. These three datasets are cross-lingual KGs extracted from DBpedia: French-English (FR-EN), Chinese-English (ZH-EN), and Japanese English (JA-EN). The number of entities in each KG is between 15K and 20K, most of which (15K) are pre-aligned.
    \item DBP-WD and DBP-YG from DWY100K~\cite{DBLP:conf/ijcai/SunHZQ18}. These two datasets are mono-lingual (English) datasets. The KGs of DBP-WD are extracted from DBpedia and Wikidata, while KGs of DBP-YG are extracted from DBpedia and Yago. Each KG contains 100K entities which are all pre-aligned.
    \item FB-DBP~\cite{zeng2022entity} is constructed for studying the division of large-scale EA. The KGs were extracted from Freebase and DBPedia. Each KG has over 2 million entities which are all pre-aligned.
\end{itemize}

\noindent
Following previous works~\cite{DBLP:journals/pvldb/SunZHWCAL20,DBLP:journals/pvldb/GeLCZG21}, we use 20\% pre-aligned mappings as training data, while the remaining mappings are used for test.
Refer to the original papers for statistics of the datasets.

\subsection{Details for Reproducibility}\label{sec:details_imp}
Some details in our implementation include:
(1) The numbers of subtasks for the datasets are set as: DBP15K: 5; DWY100K: 10; and FB-DBP (2M): 200.
(2) Both RREA and GCN-Align only exploit the KG structure because it is a more challenging setting in EA. The supervised training mode has 250 epochs, while the semi-supervised training has 5 iterations with 50 epochs in each iteration.
(3) I-SBP and our framework \divea are both performed in an iterative way for 5 iterations.
(4) Hyperparameters of our framework are set as: $\alpha=0.9$, $\beta=1.0$, $K=10$, $\gamma=2.0$.
(5) In evaluation, we match each unmatched source entity with all counterpart candidates in a subtask as in LargeEA~\cite{DBLP:journals/pvldb/GeLCZG21}, and make the evaluation consistent with all methods.~\footnote{We notice only ground-truth counterparts are used as candidates for evaluation in the source code of SBP: https://github.com/DexterZeng/LIME. This does not make sense. We modify it in this work.
}
(6) We run the experiments on one GPU server, which is configured with an Intel(R) Xeon(R) Gold 6128 3.40GHz CPU, 128GB memory, 3 NVIDIA GeForce GTX 2080Ti GPU and Ubuntu 20.04 OS.
(7) We release our codes, running scripts with parameter settings, and the used data at \url{https://github.com/uqbingliu/DivEA} for reproducing this work.

\section{Results}

\begin{table*}
    \centering
    \caption{Overall performance of \divea and CPS variants. Bold indicates best; all differences between \divea and CPS variants on DBP15K and DWY100K datasets are statistically significant (p < 0.05);}
    \vspace{-8pt}
    \scalebox{0.76}[0.76]{
    \begin{tabular}{c|c|ccc|ccc|ccc|ccc|ccc|ccc}
        \hline
        \multirow{2}{*}{Method} & \multirow{2}{*}{EA model} & \multicolumn{3}{c|}{FR-EN (15K)} & \multicolumn{3}{c|}{JA-EN (15K)}  & \multicolumn{3}{c|}{ZH-EN (15K)}  & \multicolumn{3}{c|}{DBP-WD (100K)} & \multicolumn{3}{c|}{DBP-YG (100K)} & \multicolumn{3}{c}{FB-DBP (2M)} \\
        & & H@1 & H@5 & MRR & H@1 & H@5 & MRR & H@1 & H@5 & MRR & H@1 & H@5 & MRR & H@1 & H@5 & MRR & H@1 & H@5 & MRR\\
        \hline
        CPS (sup) & \multirow{3}{*}{GCN-Align} & 0.151&0.396&0.263&0.106&0.276&0.184&0.178&0.405&0.282&0.114&0.222&0.162&0.101&0.174&0.134 &0.000&0.000&0.000 \\
        CPS (semi) & & 0.274&0.478&0.367&0.185&0.323&0.251&0.220&0.369&0.291&0.239&0.335&0.283&0.168&0.231&0.198 & 0.000&0.000&0.000 \\
        % \hline
        \divea & & \textbf{0.396}&\textbf{0.642}&\textbf{0.504}&\textbf{0.404}&\textbf{0.641}&\textbf{0.509}&\textbf{0.374}&\textbf{0.602}&\textbf{0.474}& \textbf{0.399}&\textbf{0.571}&\textbf{0.477}& \textbf{0.552}&\textbf{0.698}&\textbf{0.619} &\textbf{0.051}&\textbf{0.106}&\textbf{0.08} \\
        \hline
        \hline
        CPS (sup) & \multirow{3}{*}{RREA} &  0.419&0.631&0.514&0.329&0.507&0.409&0.436&0.628&0.522&0.337&0.445&0.385&0.320&0.416&0.363 & 0.043&0.080&0.062 \\
        CPS (semi) & & 0.516&0.682&0.590&0.385&0.525&0.448&0.512&0.666&0.582&0.389&0.470&0.426&0.357&0.434&0.392 & 0.056&0.089&0.073\\
        % \hline
        \divea & & \textbf{0.645}&\textbf{0.795}&\textbf{0.711}&\textbf{0.628}&\textbf{0.787}&\textbf{0.698}&\textbf{0.633}&\textbf{0.775}&\textbf{0.696}&\textbf{0.618}&\textbf{0.728}&\textbf{0.668}&\textbf{0.742}&\textbf{0.827}&\textbf{0.781 }&\textbf{ 0.163}&\textbf{0.24}&\textbf{0.202} \\
        \hline
    \end{tabular}
    }
    \label{tab:overperf_cps}
\end{table*}

\begin{table*}
    \centering
    \caption{Overall performance of \divea and SBP variants. Bold indicates best; all differences between \divea and SBP variants on DBP15K and DWY100K datasets are statistically significant (p < 0.05); we compare \divea with SBP variants separately because they build larger subtasks than CPS.}
    \vspace{-8pt}
    \scalebox{0.76}[0.76]{
    \begin{tabular}{c|c|ccc|ccc|ccc|ccc|ccc|ccc}
        \hline
        \multirow{2}{*}{Method} & \multirow{2}{*}{EA model} & \multicolumn{3}{c|}{FR-EN (15K)} & \multicolumn{3}{c|}{JA-EN (15K)}  & \multicolumn{3}{c|}{ZH-EN (15K)}  & \multicolumn{3}{c|}{DBP-WD (100K)} & \multicolumn{3}{c|}{DBP-YG (100K)} & \multicolumn{3}{c}{FB-DBP (2M)} \\
        & & H@1 & H@5 & MRR & H@1 & H@5 & MRR & H@1 & H@5 & MRR & H@1 & H@5 & MRR & H@1 & H@5 & MRR & H@1 & H@5 & MRR\\
        \hline
        SBP (sup) & \multirow{4}{*}{GCN-Align} &  0.163&0.426&0.284&0.122&0.336&0.222&0.182&0.439&0.301&0.085&0.166&0.12&0.111&0.198&0.150& 0.000& 0.000& 0.000\\
        SBP (semi) & & 0.288&0.511&0.391&0.249&0.452&0.347&0.289&0.497&0.388&0.142&0.193&0.165&0.173&0.231&0.200& 0.005&0.011&0.008\\
        I-SBP & & 0.175&0.372&0.267&0.160&0.386&0.265&0.210&0.452&0.323&0.170&0.247&0.205&0.473&0.64&0.548&0.000& 0.000& 0.000 \\
        % \hline
        \divea & &\textbf{0.402}&\textbf{0.678}&\textbf{0.525}&\textbf{0.413}&\textbf{0.669}&\textbf{0.527}&\textbf{0.387}&\textbf{0.641}&\textbf{0.501}&\textbf{0.481}&\textbf{0.690}&\textbf{0.575}&\textbf{0.645}&\textbf{0.819}&\textbf{0.724}& \textbf{0.071}&\textbf{0.15}&\textbf{0.112} \\
        \hline
        \hline
        SBP (sup) & \multirow{4}{*}{RREA} & 0.475&0.721&0.583&0.419&0.667&0.529&0.484&0.704&0.584&0.506&0.667&0.578&0.600&0.727&0.657 &0.070&0.139&0.106 \\
        SBP (semi) & & 0.575&0.762&0.659&0.481&0.689&0.575&0.566&0.752&0.650&0.575&0.697&0.631&0.648&0.749&0.693&0.095&0.159&0.128 \\
        I-SBP & & 0.508&0.730&0.608&0.446&0.676&0.549&0.517&0.728&0.614&0.638&0.781&0.704&0.764&0.889&0.82&0.120&0.233&0.172 \\
        % \hline
        \divea & & \textbf{0.655}&\textbf{0.841}&\textbf{0.736}&\textbf{0.647}&\textbf{0.830}&\textbf{0.728}&\textbf{0.646}&\textbf{0.819}&\textbf{0.723}&\textbf{0.701}&\textbf{0.826}&\textbf{0.757}&\textbf{0.816}&\textbf{0.913}&\textbf{0.86} & \textbf{0.199}&\textbf{0.298}&\textbf{0.248} \\
        \hline
    \end{tabular}
    }
    \label{tab:overperf_sbp}
\end{table*}

%\subsection{RQ1: Overall Performance}
\subsection{Overall Effectiveness}

Tables~\ref{tab:overperf_cps} and~\ref{tab:overperf_sbp} report the overall effectiveness of \divea and the baselines evaluated with two neural EA models on 6 datasets.
As mentioned, we compare \divea with CPS and SBP separately because the sizes of the subtasks they build are not comparable.
The experiments on 5 datasets -- FR-EN, JA-EN, ZH-EN, DBP-WD, and DBP-YG  -- are run for 5 times for significance testing.
The results may be different from the reported ones in the original papers because of different settings (see Sec.~\ref{sec:details_imp}).
In Table~\ref{tab:overperf_cps}, we compare \divea with CPS variants.
Our \divea can make the two neural EA models achieve much higher effectiveness than both variants of CPS on all datasets.
CPS does not have a specific module for counterpart discovery, and it does not consider the characteristics of either the EA task or the EA models in terms of building context graphs.
Conversely, our \divea solves the two sub-objectives -- high coverage of potential mappings and informative context graphs -- with tailored methods considering the specificity of EA.

In Table~\ref{tab:overperf_sbp}, we compare \divea with three SBP variants; we can observe that:
(1) \divea largely outperforms all the variants of SBP, across all the datasets.
Although they implement improvements over CPS, SBP variants inherit the core method of CPS, and are still limited by CPS' drawbacks. Our method shows that designing a framework alternative to CPS is a more promising direction.
(2) I-SBP has better effectiveness than SBP (sup) in most cases. However, its advantage over SBP (semi), which trains EA models in the same way, is not consistent. It can achieve improvements over SBP (semi) on the 100K datasets and 2M dataset, but has worse effectiveness on small datasets, e.g. FR-EN, JA-EN, ZH-EN.
This suggests that I-SBP, which enhances SBP by enriching the seed mappings with pseudo mappings, has limited generalisability.

By comparing CPS and SBP variants across Table~\ref{tab:overperf_cps} and \ref{tab:overperf_sbp}, we find the best SBP variant is usually better than the best CPS variant with the same subtask number. But the subtask sizes of SBP are not controllable and larger than CPS. In contrast, \divea can flexibly control the subtask size while achieving better effectiveness.

In both tables, it can be observed that CPS (semi) and SBP (semi), which both train EA models in a semi-supervised manner, outperform their corresponding supervised version CPS (sup) and SBP (semi) at the same learning cost.
Therefore, semi-supervised training is recommended, and it can be used to assist EA division, as in our framework.
In addition, the state-of-the-art EA model RREA has better effectiveness than GCN-Align on the divided subtasks, a finding consistent within the whole task scenario. Both models have poor effectiveness on the 2M dataset (i.e. FB-DBP).

In the following sections, we just compare our methods with the best variants of CPS and SBP -- CPS (semi) and I-SBP. CPS denotes CPS (semi) if not specially clarified.

\subsection{Coverage of Counterpart Discovery}
The coverage of potential mappings in subtasks is essential for EA division and determines the upper bound of final EA effectiveness. We examine different EA division methods and report the recalls of ground-truth counterparts in their divided subtasks in Table~\ref{tab:recall_counterpart}. I-SBP and \divea run with RREA. We make the following observations:
(1) CPS has decent coverages of potential mappings on the DBP15K datasets, but poor coverages on larger datasets, i.e. $>50\%$ of \divea on DBP-WD and DBP-YG (100K),  and $<50\%$ of \divea on FB-DBP (2M).
(2) SBP outperforms CPS by enhancing it with the bidirectional partition strategy.  I-SBP brings the coverage to a new level by enriching the seed mappings with the EA model iteratively.
(3) \divea can achieve higher coverage than CPS, SBP and I-SBP on most datasets. By comparing the last and second lines of the table, we can see that a larger subtask size leads to higher coverage of potential mappings in \divea.

\begin{table}
    \caption{Recall of ground-truth counterparts in subtasks}
    \vspace{-8pt}
    \scalebox{0.9}[0.9]{
\begin{tabular}{c|c|c|c|c|c|c}
    \hline
    Method & FR-EN & JA-EN & ZH-EN & DBP-WD & DBP-YG & FB-DBP \\
    \hline
    CPS & 0.817&0.718&0.826 & 0.542&0.486&0.237 \\
    \divea &\textbf{0.881}&\textbf{0.892}&\textbf{0.880}&\textbf{0.830}&\textbf{0.893}&\textbf{0.507} \\
    \hline
    % SBP (subtask) & 0.837&0.768&0.845&0.633&0.646 \\
    SBP & 0.930&0.942&0.943&0.819&0.824& 0.426\\
    % I-SBP (subtask) & 0.883&0.899&0.906&0.878&0.942 \\
    I-SBP & 0.960&0.957&0.960&0.947&0.982&0.502 \\
    \divea & \textbf{0.978}&\textbf{0.979}&\textbf{0.97}&\textbf{0.954}&\textbf{0.994}& \textbf{0.684} \\
    \hline
\end{tabular}
}
    \label{tab:recall_counterpart}
\end{table}

% \subsection{RQ6: Counterpart Discovery}

\begin{figure}
    \vspace{-10pt}
    \includegraphics[width=8.7cm]{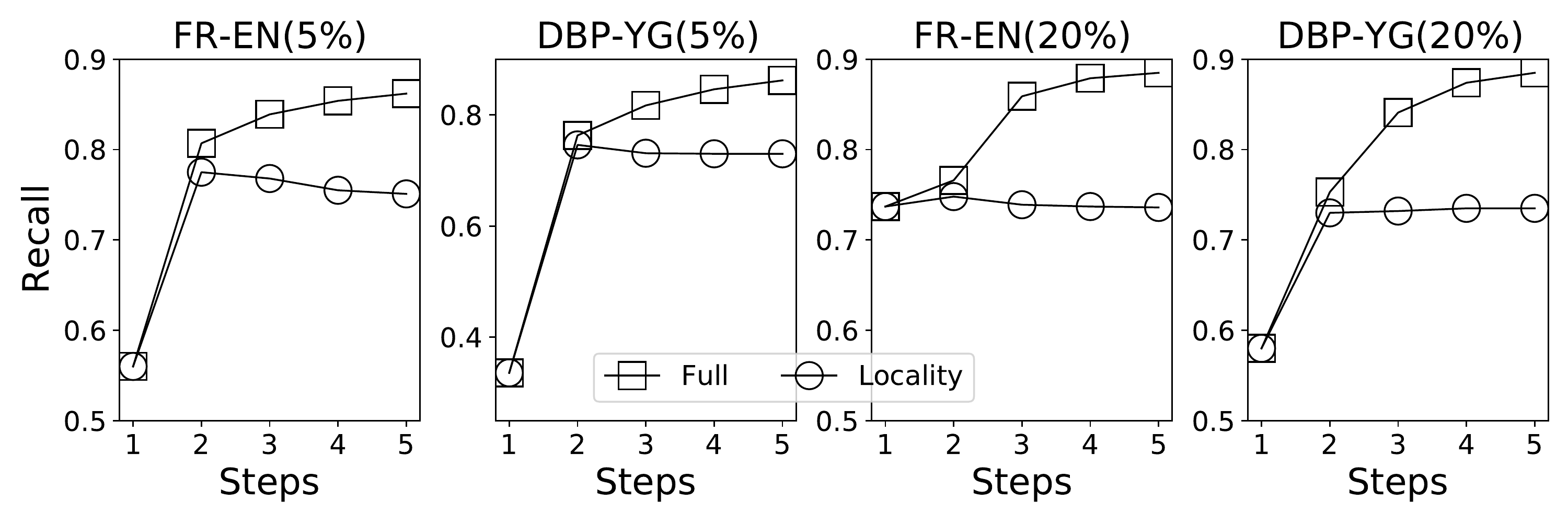}
    \vspace{-18pt}
    \caption{Examining our progressive counterpart discovery.}
    \label{fig:progressive_counterdiscovery}
    \vspace{-10pt}
\end{figure}

Furthermore, we examine our counterpart discovery to answer the following questions: (1) Is the locality principle effective in counterpart discovery? (2) Can EA model enhance the locality feature-based method by enriching the seed mappings? (3) What is the effect of the EA similarity-based feature?
To answer these questions, we run our full counterpart discovery and its variant, which only uses the locality for scoring counterpart candidates but still exploits the EA model to enrich the seed mappings in the iterative process.
In Fig.~\ref{fig:progressive_counterdiscovery}, we draw the recall of ground-truth counterparts in the iterative processes of both variants on two datasets with 5\% and 20\% training data.
It can be observed that: (1) The starting point is purely based on the locality principle without any assisting information from the EA model. The achieved performance is fair. This verifies the usefulness of locality for counterpart discovery. (2) The locality-based method can get obvious improvement in the second iteration. This is because the EA model predicts new mappings to enrich the set of seed mappings. Then, however, the coverage of the locality-based method stops increasing and even have a slight decrease.
(3) Our full method brings the coverage to a new level, even when there is only 5\% training data. This success comes from the introduction of the similarity-based feature, which makes the promising counterparts found in previous iterations still kept in the candidates.

\subsection{Informativeness of Context Builder}
%\subsection{RQ3: Informativeness of Context Builder}
\begin{figure}[t]
    \centering
    \includegraphics[width=8.7cm]{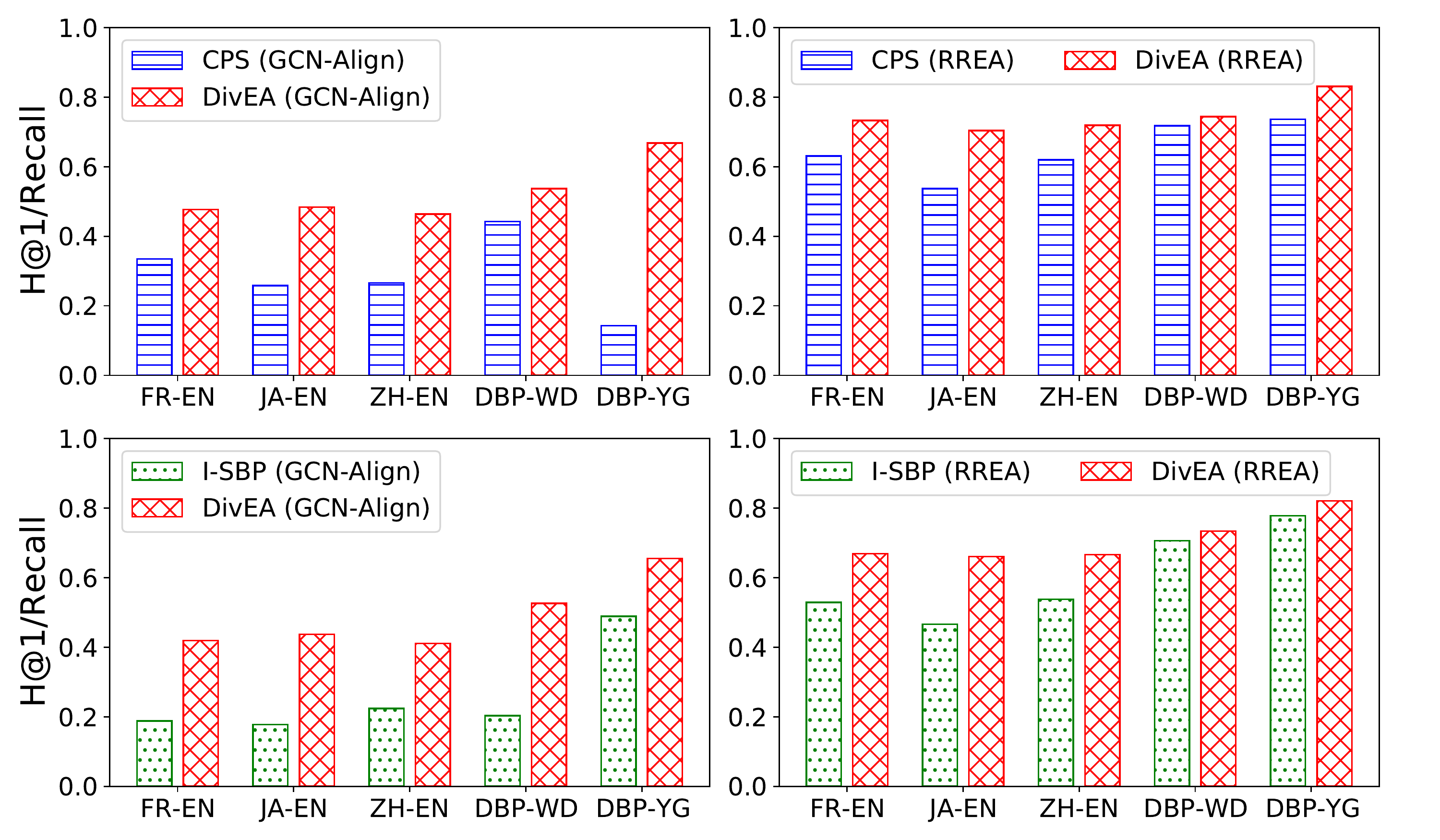}
    \vspace{-18pt}
    \caption{Informativeness of different context builders.}
    \label{fig:ctx_builder}
    \vspace{-10pt}
\end{figure}

\begin{figure}[t]
    \centering
    \includegraphics[width=8.7cm]{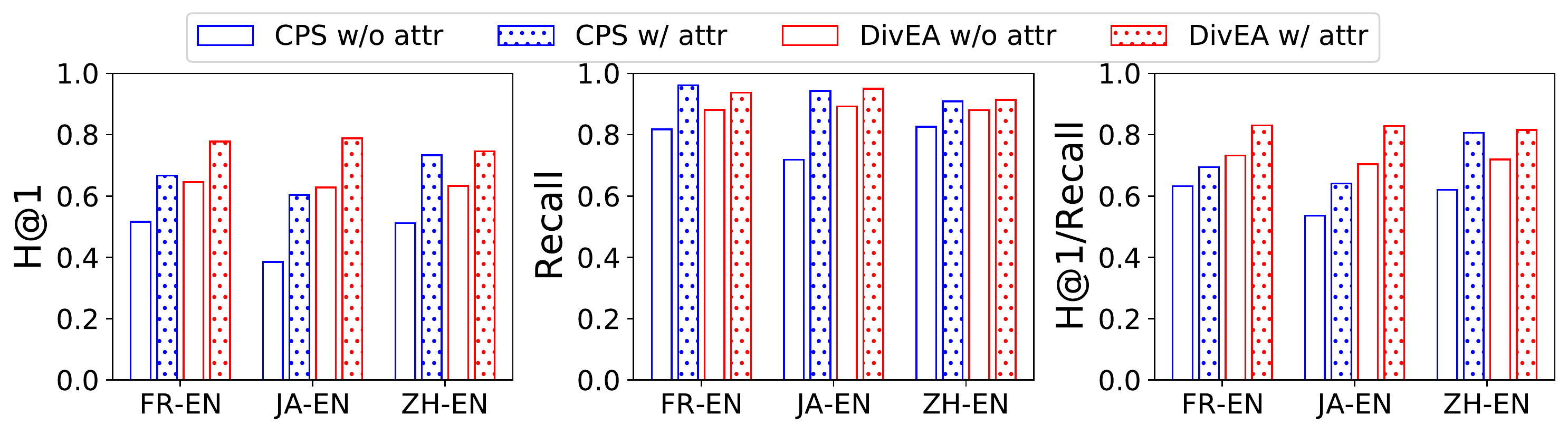}
    \vspace{-18pt}
    \caption{Incorporating attribute information.}
    \label{fig:attr}
    \vspace{-16pt}
\end{figure}

\begin{figure}[t]
    \includegraphics[width=8.7cm]{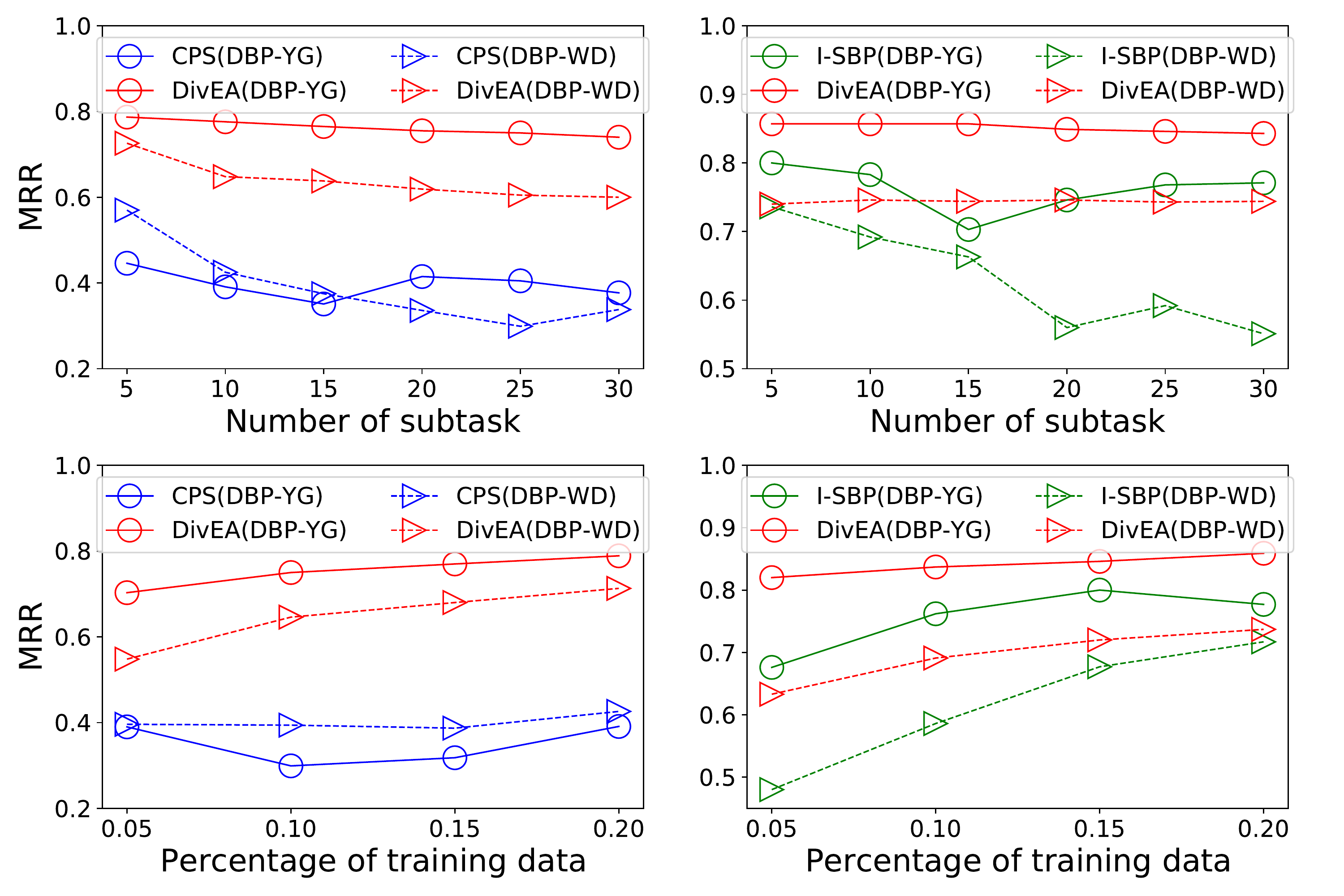}
    \vspace{-16pt}
    \caption{Robustness w.r.t. subtask number and the percentage of traning data.}
    \label{fig:robustness_partN}
    \vspace{-16pt}
\end{figure}

The informativeness of contexts is reflected by the final EA performance, which is also affected by the coverage of potential mappings. However, it is hard to control the coverage of different EA division methods at the same level.
To compare different context builders, we measure the informativeness of context with the ratio of $H@1/Recall$, which is the proportion of mappings that can be recognised by the EA model from these found by counterpart discovery.
The upper two plots in Fig.~\ref{fig:ctx_builder} show the comparison of \divea and CPS (semi) and I-SBP, which are the best variants of the two existing methods.
Each plot is evaluated with one EA model on 5 datasets.
\divea can create more informative context graphs for both neural EA models than CPS (semi) on all the datasets.
Similarly, we show the comparison between \divea and I-SBP in the lower two plots. \divea also outperforms I-SBP in building informative context in different settings.
By comparing the upper two plots and the lower two plots separately, we can see that the better EA model can capture more information than the worse EA model from the same context builder.

\subsection{Incorporating Attributes in KGs}

Ge et al.~\cite{DBLP:journals/pvldb/GeLCZG21} propose to exploit the attribute information, i.e. entity name, to improve CPS. They generate some pseudo mappings based on entity names to augment the seed mappings before performing EA division.
We check if this strategy can also be used to improve \divea.
To this end, we take the same strategy to augment seed mappings for both CPS and \divea on three datasets from DBP15K.
In Fig.~\ref{fig:attr}, we compare CPS and \divea versions with and without using attributes from three aspects: overall performance, coverage of potential mappings, and informativeness of context graphs, which are measured with H@1, Recall, and $H@1/Recall$ respectively.
In the first plot, we can see that both CPS and \divea can get improvement by data augmentation. Our \divea outperforms CPS in both scenarios.
After data augmentation, CPS in fact has very close coverage of potential mappings as \divea, as shown in the second plot. However, it still has an obvious gap from \divea w.r.t. the informativeness of context graphs, as shown in the third plot.

\subsection{Robustness}

To investigate the robustness of \divea, we examine the effect of subtask number and percent of training data on \divea and the baselines, with RREA on 100K datasets DWY-WD and DWY-YG. In the upper plots of Fig.~\ref{fig:robustness_partN}, we compare \divea with the two baselines separately across different numbers of subtasks. It is shown that: (1) the increased subtask number tends to make the EA division methods decrease in effectiveness. (2) Compared with CPS and I-SBP, \divea performs much better no matter the number of subtasks. Also, \divea is less sensitive to the change of subtask number.

\noindent
The lower two plots in Fig.~\ref{fig:robustness_partN} show the comparison between \divea and the two baselines w.r.t. different percentages of training data. We can observe that: (1) Our \divea performs better than CPS and I-SBP consistently. (2) CPS does not achieve better performance when the training data are increased. We think the reason is that CPS has poor coverage of potential mappings (around 0.5 shown in Table~\ref{tab:recall_counterpart}), which is independent of the training data.

\noindent
To conclude, our \divea is robust w.r.t. subtask number and the amount of training data.

\subsection{Gains and Costs of EA Division}

\begin{figure}[t!]
    \includegraphics[width=8.7cm]{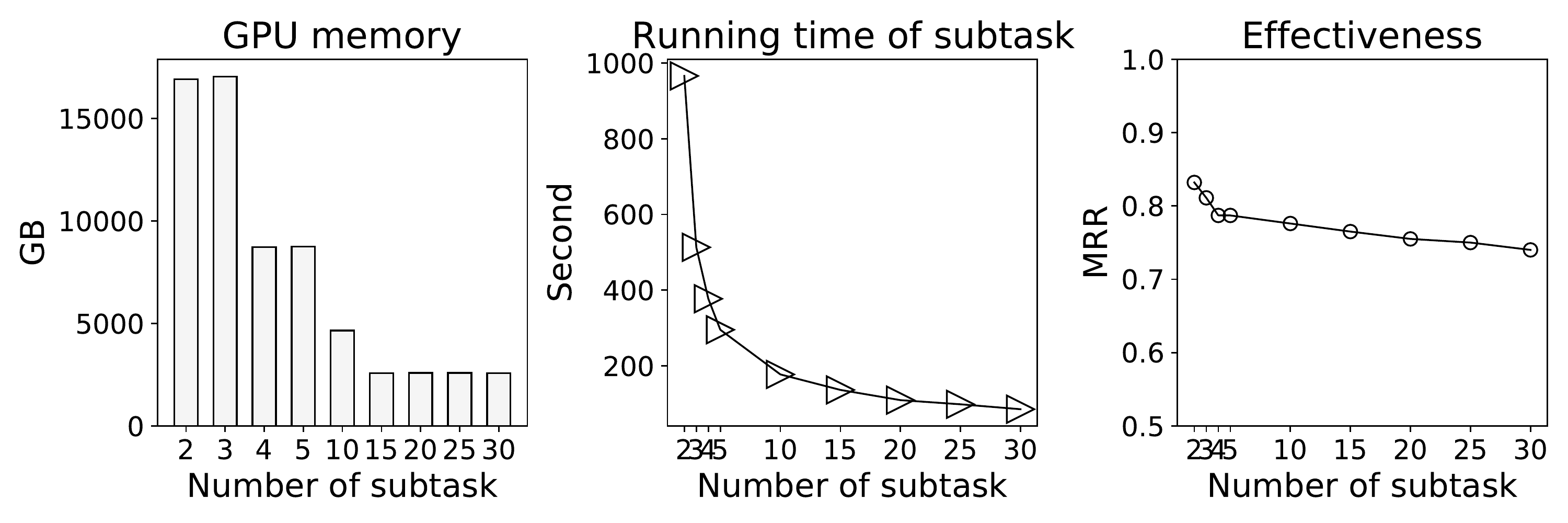}
    \vspace{-20pt}
    \caption{Gains and costs of EA task division.}
    \label{fig:gains_n_costs}
    \vspace{-10pt}
\end{figure}

\begin{figure}
    \includegraphics[width=8.7cm]{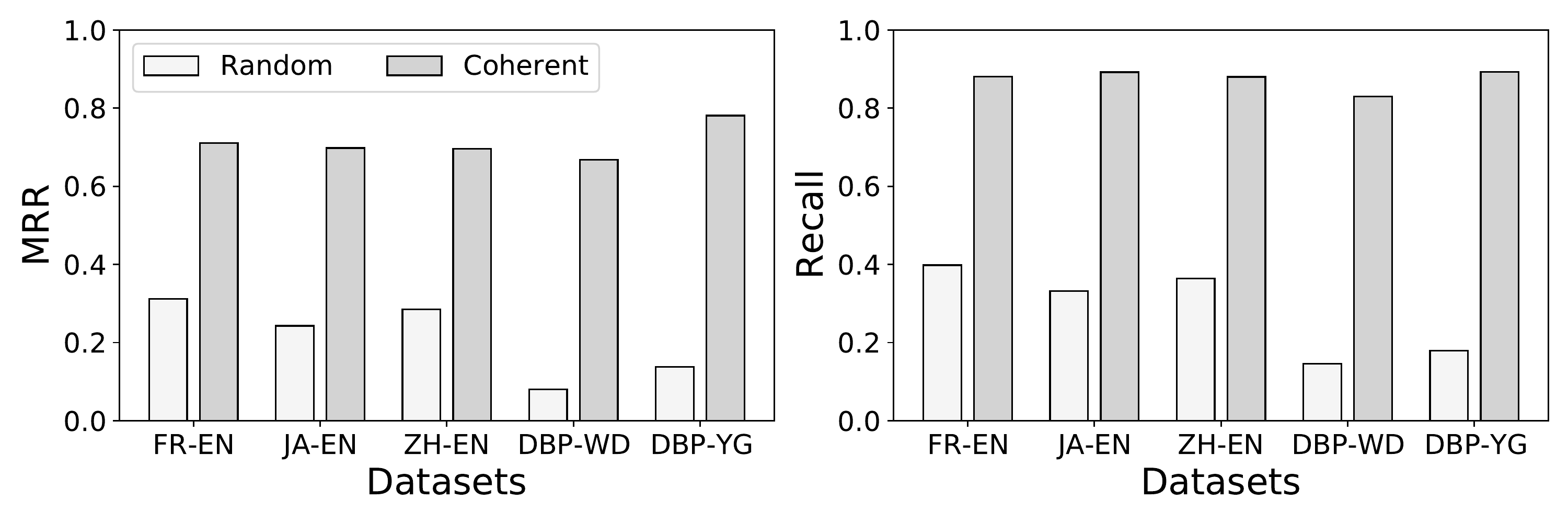}
    \vspace{-20pt}
    \caption{Necessity of coherence in dividing unmatched source entities.}
    \label{fig:random_vs_coherence}
    \vspace{-8pt}
\end{figure}

\begin{table}
    \caption{Time cost (seconds) of EA division -- CPS vs \divea. sub.: subtask; par.: parallel.     \vspace{-8pt}}
    \begin{tabular}{c|c|c|c|c}
    \hline
    Dataset (sub. num.) & EA & CPS & \divea & \divea (par.) \\
    \hline
    DBP-YG (10) & 1490 & 23 & 335 & 36 \\
    FB-DBP (200) & 42961 & 7885 & 45347 & 404 \\
    \hline
    \end{tabular}
    \label{tab:time_vs_cps}
\end{table}

\begin{table}
    \caption{Time cost (seconds) of EA division -- I-SBP vs \divea. sub.:  subtask; par.: parallel. \vspace{-8pt}}
    \begin{tabular}{c|c|c|c|c}
    \hline
    Dataset (sub. num.) & EA & I-SBP & \divea & \divea (par.) \\
    \hline
    DBP-YG (10) & 1820 & 336 & 398 & 43 \\
    FB-DBP (200) & 59999 & 98412 & 52944 & 428 \\
    \hline
    \end{tabular}
    \label{tab:time_vs_isbp}
    \vspace{-6pt}
\end{table}

To study the benefits and costs of EA division, we execute \divea with RREA on DBP-YG (100K) with different subtask numbers.
In Fig.~\ref{fig:gains_n_costs}, we show the change in GPU memory usage, run time of each subtask, and effectiveness w.r.t. different subtask numbers.
The increase in subtask number can greatly reduce the requirement in GPU memory and run time for each subtask. Both quantities decrease sharply at the beginning but tend to be more stable afterwards.
However, the overall effectiveness worsens with the increase in subtask number.
Therefore, the number of subtasks represents a trade-off between lower memory utilisation and parallelism support on one hand, and reduced effectiveness on the other.

The extra time spent in EA division is another concern. We report the extra time spent by \divea and two baselines on datasets of different sizes: DBP-YG (100K) and FB-DBP (2M).
(1) Table~\ref{tab:time_vs_cps} shows \divea spends much more time than CPS no matter the size of datasets.
(2) Table~\ref{tab:time_vs_isbp} shows \divea has comparable time consumption with I-SBP on DBP-YG (100K), but  it is much faster on FB-DBP (2M).
(3) Most of \divea's workload occurs in the subtasks. If subtasks are run in parallel, the extra time consumption for division can also be greatly reduced. Tables~\ref{tab:time_vs_cps} and \ref{tab:time_vs_isbp} report the time cost of \divea in parallel mode.
Note that CPS and SBP do not have this feature.
(4) Without considering the parallelism, the time for EA division in \divea is comparable to that of training EA models.

\subsection{Necessity of Coherence in Unmatched Source Entity Division}

To show the necessity of coherence for dividing unmatched source entities, we replace the coherent division with random division in \divea. In particular, we start by splitting all source entities randomly. Then, in each split, the unmatched entities form one group, while the anchors present are used to look for the target counterparts.
In Fig.~\ref{fig:random_vs_coherence}, we compare random division with our coherent division on five datasets and two metrics -- overall effectiveness and recall of counterpart discovery. The neural EA model used is RREA.
As expected, a random division causes poor overall performance and coverage of potential mappings.
Therefore, coherence in source entity division is necessary for our framework.

\section{Conclusion}

Entity Alignment is a key task for Knowledge Graph fusion. Existing neural EA models cannot run on large-scale KGs due to critical memory issues.
To make the neural EA models be able to run one large-scale KGs, few previous works investigated dividing a large-scale EA task into small subtasks.
However, they can achieve limited EA effectiveness because they suffer from low coverage of potential mappings, insufficient evidence of context graphs, and varying subtask sizes. 
In this work, we propose a new framework called \divea, which decouples unmatched source entity division, counterpart discovery, and context building.  
To achieve high coverage of potential mappings, we exploit the locality principle of the EA task and current EA model.
To build informative context graphs, we design an evidence passing mechanism which can quantify the informativeness of context graphs.
In our framework, the subtask size can be controlled flexibly. 
Extensive experiments show that \divea outperforms the existing methods significantly across different neural EA models. \divea obtains higher coverage of potential mappings, builds more informative context graphs, and is robust w.r.t. subtask number and amount of training data.

% In future, we will investigate how to improve EA performance on larger datasets than those considered in our empirical experiments.
% We note that EA performance on the 2M dataset FB-DBP is poor.
% A reason for this is that \divea still has large space for improvement, e.g. coverage of potential mappings. 
% Another reason is the state-of-the-art neural EA models proposed for small datasets cannot fit the challenges present in large-scale datasets. With the assistance of EA division frameworks, like \divea, researchers have the chance to design neural EA models for large-scale KGs.


\begin{acks}
This research is supported by the National Key Research and Development Program of China No. 2020AAA0109400, the Shenyang Science and Technology Plan Fund (No. 21-102-0-09), and the Australian Research Council (No. DE210100160 and DP200103650).
\end{acks}

\bibliographystyle{ACM-Reference-Format}
\balance
\bibliography{bibfile}

\appendix
% \input{sections/note.tex}
% placeholder

\end{document}